\documentclass[pdflatex,sn-basic,iicol]{sn-jnl}

\usepackage{graphicx}%
\usepackage{multirow}%
\usepackage{amsmath,amssymb,amsfonts}%
\usepackage{amsthm}%
\usepackage[title]{appendix}%
\usepackage{xcolor}%
\usepackage{textcomp}%
\usepackage{manyfoot}%
\usepackage{booktabs}%
\usepackage{algorithm}%
\usepackage{algorithmicx}%
\usepackage{algpseudocode}%
\usepackage{listings}%

\usepackage{enumitem}
\usepackage{multirow}
\usepackage{makecell}

\theoremstyle{thmstyleone}%

\theoremstyle{thmstyletwo}%

\theoremstyle{thmstylethree}%

\begin{document}

\title[Article Title]{OmniRaindrop: A Comprehensive Benchmark for Dual-Focused Raindrop Removal under Day and Night Scenes}

\author[1]{\fnm{Zhiqiang} \sur{Teng}}
\equalcont{These authors contributed equally to this work. \textsuperscript{\textdaggerdbl}Project leader.}
\author[2]{\fnm{Yeying} \sur{Jin}}
\equalcont{These authors contributed equally to this work. \textsuperscript{\textdaggerdbl}Project leader.}
\author[2]{\fnm{Zifeng} \sur{Yuan}}
\author[2]{\fnm{Tingting} \sur{Chen}}
\author[2]{\fnm{Beibei} \sur{Lin}\textsuperscript{\textdaggerdbl}}
\author[3]{\fnm{Xin} \sur{Li}}
\author*[1]{\fnm{Shunli} \sur{Zhang}}\email{slzhang@bjtu.edu.cn}

\affil*[1]{%
  \orgname{Beijing Jiaotong University},
  \orgaddress{
    \city{Beijing},
    \postcode{100044},
    \country{China}
  }
}

\affil[2]{%
  \orgname{National University of Singapore},
  \orgaddress{
    \city{Singapore},
    \postcode{119077},
    \country{Singapore}
  }
}

\affil[3]{%
  \orgname{University of Science and Technology of China},
  \orgaddress{
    \city{Hefei},
    \postcode{230026},
    \country{China}
  }
}


\abstract{
Adherent raindrops on camera lenses are a pervasive source of real-world image degradation, undermining outdoor visual perception for downstream applications such as robotics and 3D scene reconstruction.
However, most existing raindrop datasets and methods target only background-focused 2D images, overlooking the equally common raindrop-focused case and the impact of raindrops on 3D scene reconstruction.
Moreover, they are almost exclusively daytime, lacking the complex artificial lighting and reflections of nighttime raindrop scenes.
To address these issues, we introduce \textbf{OmniRaindrop}, a real-world dataset and benchmark that jointly evaluates 2D raindrop removal and 3D scene reconstruction across raindrop-focused and background-focused captures, under both daytime and nighttime conditions.
OmniRaindrop provides spatially aligned raindrop-focused, background-focused, and rain-free ground-truth image sets: a 2D subset of 15,186 aligned pairs/triplets and a 3D subset of 40 real-world scenes evenly split between day and night, totaling 3,822 multi-view images.
On top of this data, the benchmark traces how raindrop degradation propagates from 2D restoration to 3D reconstruction by testing representative restoration methods and their effects on pose estimation, point cloud initialization, and 3DGS optimization.
Comprehensive experiments reveal that raindrop-focused scenes are substantially harder than background-focused ones, and that errors in 2D restoration and pose/point cloud initialization cascade into degraded 3D reconstruction, pointing to clear directions for more robust 2D restoration and 3DGS design.
Building on the dataset, we have organized two consecutive NTIRE challenges (2025 and 2026) on Day and Night Raindrop Removal for Dual-Focused Images.}





\maketitle

\section{Introduction}\label{sec1}

%
Adherent raindrops on camera lenses are a common source of degradation in outdoor visual perception systems~\cite{hao2019learning, yang2020raindrop, you2015adherent, chen2025towards}.
Acting as miniature convex lenses, they introduce complex optical distortions, localized light refractions, and severe occlusions~\cite{guo2020joint, li2024dual, luo2020weakly, liu2021learning}. 
These artifacts can compromise the reliability of downstream applications. In autonomous driving, lens raindrops can obscure or distort visual evidence needed for detecting and localizing pedestrians, vehicles, and scene structures, thereby increasing safety risks~\cite{brophy2023review, hamzeh2021review, qiu2024visual}.
 In robotic navigation, raindrop occlusions disrupt feature tracking in SLAM, causing mapping failures and trajectory deviations~\cite{steffens2021robustness}. Beyond raindrop-specific degradation, visual SLAM and localization remain challenging under large seasonal, weather, and illumination variations~\cite{wenzel20254seasons}. In surveillance reconstruction, raindrop ghosting can obscure or distort critical cues, such as facial details and license plate information, potentially degrading the reliability of reconstructed 3D models for forensic analysis~\cite{lee2023estimation, you2015adherent, soboleva2021raindrops, yang2026rethinking}.

\begin{figure}[t!]
	\centering
	{\includegraphics[width=1.0\linewidth]{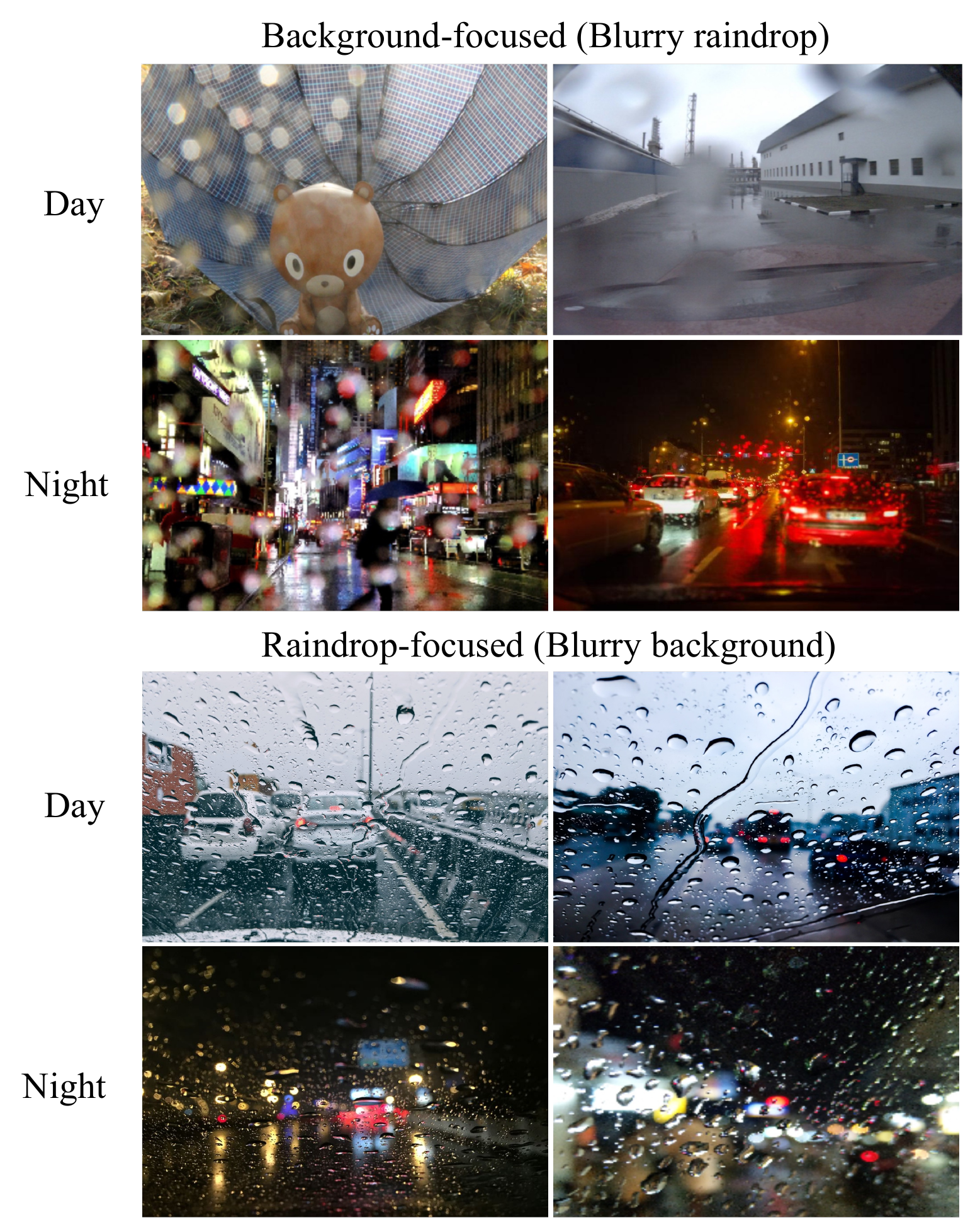}}\hfill	
	\caption{Motivation for OmniRaindrop. Top block: background-focused captures (blurry raindrops)~\cite{qian2018attentive, soboleva2021raindrops, jin2024raindrop}; bottom block: raindrop-focused captures (blurry background); within each block, the upper and lower rows show daytime and nighttime, respectively~\cite{jin2024raindrop}.}
	\label{fig:dataset}
\end{figure}

Recently, a few raindrop datasets (e.g.,~\cite{qian2018attentive,quan2021removing,porav2019can,soboleva2021raindrops}) have been proposed for single-image (2D) raindrop removal.
%
%
%
However, they operate under an idealized setup of daytime illumination and a static camera focus on the background, which leaves two major 2D limitations.
First, camera focus shifts create two practically important focus conditions (Fig.~\ref{fig:dataset}): a raindrop-focused setting that severely blurs the background, and a background-focused setting where raindrops appear as diffuse, blurred occlusions~\cite{jin2024raindrop}. As a result, methods designed for blurry raindrops over a sharp background tend to perform poorly on raindrop-focused inputs.
%
%
Second, a severe domain gap separates daytime and nighttime raindrop degradation. Unlike daytime scenes, nighttime scenes involve highly location- and illumination-dependent artifacts caused by artificial light transport, glow, glare, reflections, and increased sensor noise~\cite{lin2024nightrain, zhang2025learning}. Consequently, models trained solely on daytime data often fail to generalize to real-world nighttime scenarios.

A further limitation emerges when moving from 2D restoration to 3DGS reconstruction. Beyond 2D appearance, raindrops can affect multiple stages of the 3DGS reconstruction pipeline, a dimension that existing benchmarks largely overlook by treating adherent raindrops as a purely single-image problem.
A few 3D raindrop datasets~\cite{li2024derainnerf, liu2025deraings, qian2025weathergs} attempt to benchmark 3DGS reconstruction under raindrops, but they remain confined to daytime and background-focused settings, inheriting the same limitations as their 2D counterparts.
Moreover, their raindrops are synthetically rendered onto clear images~\cite{li2024derainnerf, liu2025deraings}, leaving a substantial domain gap from physically formed adherent raindrops.
More critically, when benchmarking 3DGS methods, these datasets take accurate camera poses and dense point cloud initialization for granted, whereas in real scenes,  adherent raindrops weaken cross-view feature correspondence, corrupt pose estimation and point cloud initialization even before 3DGS optimization begins~\cite{kwon2025r3evision, pan2025liberated, matteo20246dgs, chen2024zerogs, wang2025r3gs, ling2025lle}.

\begin{figure*}[t!]
    \centering
    \includegraphics[width=\textwidth]{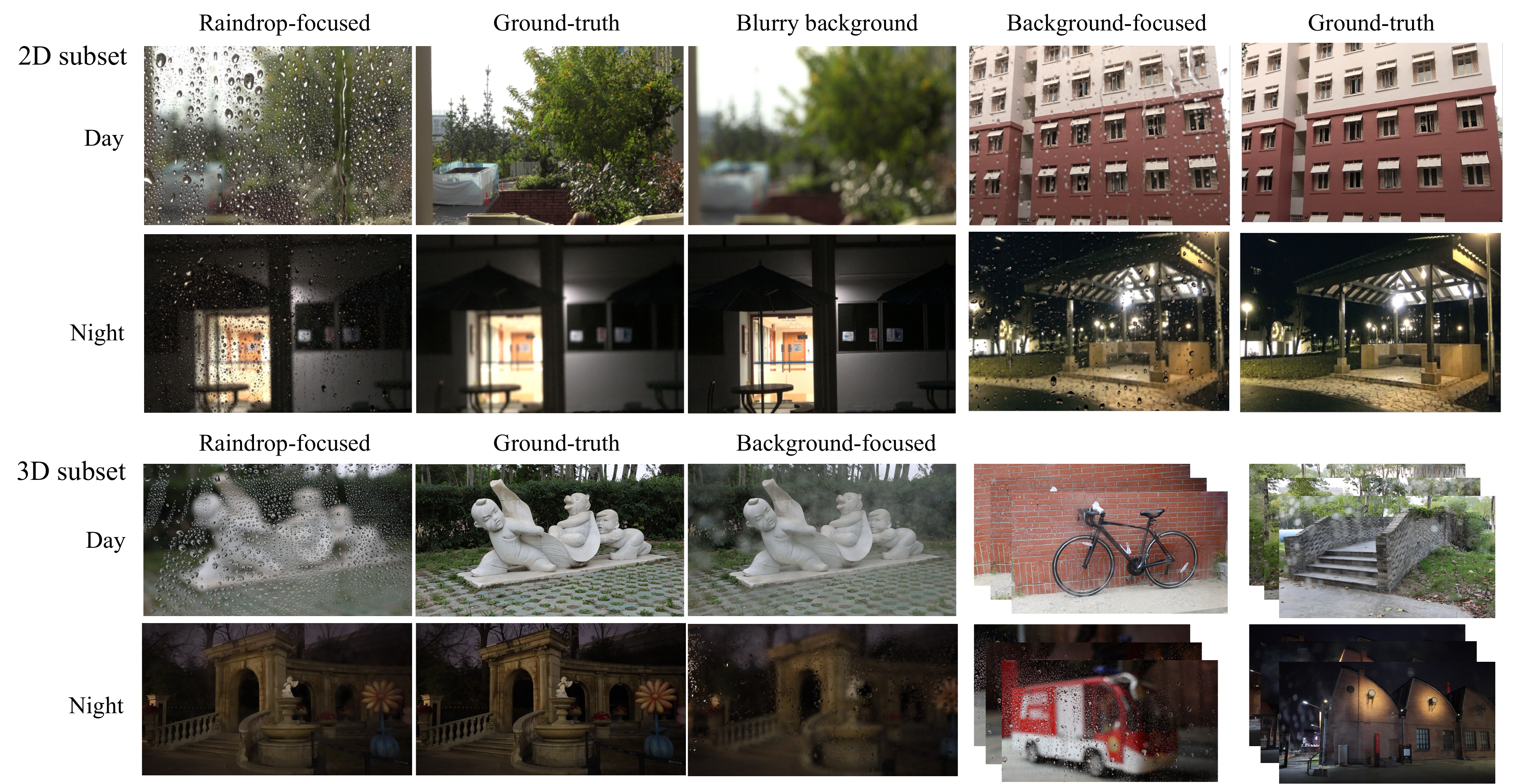}
    \caption{Illustration of the OmniRaindrop dataset. OmniRaindrop covers day/night and raindrop-/background-focused settings, providing aligned 2D pairs/triplets and real 3D multi-view scenes with rain-free references for evaluating 2D raindrop removal and 3DGS reconstruction.}
    \label{fig:fig_main}
    \vspace{-3mm}
\end{figure*}

Motivated by these observations, we propose \textbf{OmniRaindrop}, the first large-scale real-world benchmark that unifies dual-focused, day-and-night raindrop degradation across the full pipeline, from 2D restoration to 3D Gaussian reconstruction. Fig.~\ref{fig:fig_main} provides an overview of the 2D and 3D subsets of the dataset across different focus and illumination conditions.
To collect aligned real-world data, we mount a glass plate in front of a stationary camera and, by switching the focal plane, capture aligned image triplets of the same static scene: a rain-free clear reference, a raindrop-focused image with sharp raindrop structures and a defocused background, and a background-focused image with a sharp background and diffuse raindrop occlusions.
For each triplet, we compute a difference map and manually inspect the result to filter out misaligned or low-quality samples.

This pipeline covers both daytime and nighttime environments, recording the complex highlights and non-linear photometric distortions introduced by artificial nighttime light sources.
Our dataset integrates over 15,000 high-quality 2D pairs and triplets, captured by spraying water at various patterns and distances, to model diverse photometric variations and background defocus \cite{jin2024raindrop}.
We additionally acquire multi-view sequences for 40 distinct real-world 3D scenes in the same way, providing synchronized raindrop-focused, background-focused, and rain-free ground truth for a comprehensive evaluation of the entire vision pipeline.


Based on this dataset, we conduct a comprehensive benchmark across the full pipeline from 2D restoration to 3DGS reconstruction, aiming to reveal the actual bottlenecks of existing pipelines under real-world focus variations and complex illumination.

For 2D restoration, we select baselines spanning both domain-specific deraining methods and general-purpose restoration backbones.
The domain-specific group includes specialized deraining algorithms (e.g., IDT \cite{xiao2022IDT}, RainDS \cite{quan2021removing}, and RDiffusion \cite{ozdenizci2023restoring}) to test their generalization in unconstrained extreme environments.
The general-purpose group adopts state-of-the-art backbones (the Transformer-based Restormer \cite{zamir2022restormer}, Uformer \cite{wang2022uformer}, and DiT \cite{peebles2023scalable}) to examine whether modern long-range and generative modeling can overcome the structural background loss caused by focal shifts.
Through extensive quantitative testing and qualitative failure analysis of these baselines, we provide empirical guidance for developing robust raindrop removal methods.

Building on the dataset, our 3D benchmark focuses on two core stages: data processing and 3DGS evaluation.
In data processing, we discard the perfect-initialization assumption by performing camera pose estimation and point cloud initialization directly on raindrop-corrupted images, with single-image deraining integrated as a front-end module.
The processed data is then fed into various state-of-the-art 3DGS architectures for extensive evaluation.
This design lets us systematically analyze how early-stage errors (inaccurate camera poses, sparse point clouds, and degraded 2D restoration) cascade through the pipeline, providing clear directions for robust 3DGS methods under raindrop conditions.

Extensive evaluations across both dimensions yield several key insights: the bottlenecks of existing methods, the restoration paradox between single-view enhancement and multi-view consistency, and the cascading effect of raindrop degradation from 2D appearance to 3D geometry.
Our 2D analysis reveals that raindrop-focused scenes present a substantially higher barrier than background-focused ones, since existing models struggle to simultaneously eliminate sharp occlusions and recover defocused background textures.
In the 3DGS pipeline, this paradox manifests as single-view enhancements introducing multi-view inconsistencies and structural artifacts that degrade the fidelity of the final reconstruction.
Furthermore, raindrop-induced distortions severely degrade cross-view feature matching, leading to significant failures in camera pose estimation and point cloud initialization.
Collectively, these findings show how 2D photometric artifacts trigger cascading errors that eventually degrade 3D geometric consistency and rendering quality.
By exposing these vulnerabilities, OmniRaindrop provides a comprehensive benchmark for developing and evaluating robust vision systems capable of preserving geometric consistency in raindrop environments.

The main contributions of this work are:
\begin{itemize}[nosep, leftmargin=*]
\item We collect a large-scale, real-world 2D dual-focused raindrop dataset spanning daytime and nighttime, and establish a 2D benchmark over representative raindrop-removal and general restoration methods on both background-focused and raindrop-focused images.
\item We collect real-world paired 3DGS scenes under both daytime and nighttime conditions, each containing three aligned image sets (raindrop-focused, background-focused, and rain-free ground truth), enabling comprehensive 3D reconstruction evaluation across focus regimes and illumination settings.
\item We introduce the first 3DGS benchmark for raindrop-contaminated scenes, covering the complete pipeline from unconstrained, raindrop-corrupted images to the final 3D Gaussian reconstructions.
\item We comprehensively validate existing 2D raindrop removal and restoration methods and 3DGS methods on our benchmark, revealing their strengths and limitations, and providing insights into future research directions.
\end{itemize}

This paper is a substantial extension of our preliminary conference version published at ECCV 2024~\cite{jin2024raindrop}.
The major improvements are as follows.
\textbf{(1)} We extend the benchmark from 2D to 3D by collecting a new real-world 3D raindrop dataset of 40 multi-view scenes (20 daytime and 20 nighttime), each providing aligned raindrop-focused, background-focused, and rain-free triplets.
\textbf{(2)} We build the first complete 3DGS evaluation pipeline for raindrop-contaminated scenes, performing camera pose estimation and point cloud initialization directly on raindrop-corrupted images before feeding them into representative 3DGS architectures.
\textbf{(3)} We substantially expand the evaluation, additionally benchmarking modern 2D restoration backbones and 3DGS methods, and tracing how 2D restoration propagates into 3D reconstruction.
\textbf{(4)} We provide in-depth analysis, new insights, and discussion on the coupled photometric–geometric degradation introduced by real raindrops.
Driven by this benchmark, we further organized two consecutive NTIRE challenges (2025 and 2026) on Day and Night Raindrop Removal, demonstrating its widespread impact within the community~\cite{lee2025ntire, sun2026ntire}.

\section{Related Work}
\label{sec:RelatedWork}

\subsection{Raindrop Removal Datasets and Methods}
Substantial research has been dedicated to remove raindrops from single image. In terms of datasets, early benchmarks (e.g., Robot Car, RainDS, and Windshields) have significantly advanced the field \cite{porav2019can, quan2021removing, qian2018attentive, soboleva2021raindrops}; however, they are predominantly confined to daytime, background-focused scenes, lacking representation of nighttime illumination and raindrop-focused conditions \cite{jin2024raindrop}. 

To mitigate lens occlusion artifacts, single-image raindrop removal methods have been extensively studied. Early approaches, such as the attention-guided method of Qian.~\cite{qian2018attentive}, employ visual attention to localize raindrop regions and reconstruct the occluded background. UMAN~\cite{shao2021uncertainty} further improves this paradigm by introducing uncertainty-aware modeling and multiscale feature fusion. More recently, transformer-based and general-purpose restoration backbones, such as Restormer~\cite{zamir2022restormer}, Uformer~\cite{wang2022uformer}, DiT~\cite{peebles2023scalable}, and IDT~\cite{xiao2022IDT}, have shown strong restoration capability by modeling long-range dependencies and global image structures.
Recent restoration studies and surveys further show a trend toward unified models for mixed degradations and efficient image restoration~\cite{jiang2025survey,li2025diffusion,guo2025efficientderain+,wu2026learning,wang2024gridformer,fei2026generative}, but real raindrop degradation under different focus settings remains underexplored.
Moreover, these methods process each image independently and do not explicitly enforce cross-view consistency, which may lead to reconstruction artifacts when they are used as preprocessing modules for 3D reconstruction.

\subsection{3DGS Reconstruction under Raindrop Conditions}
In recent years, 3DGS has emerged as a powerful technique for scene reconstruction~\cite{kerbl20233d}. Recent 3DGS methods that incorporate geometric constraints indicate that photometric supervision alone can produce irregular Gaussian distributions, reinforcing the need to analyze geometric degradation under raindrop conditions~\cite{fang2025efficient, zhan2026rdg}.
Beyond 3DGS, related 3D vision studies have considered complex-reflectance reconstruction, non-rigid reconstruction, and neural-rendering-based point-cloud representation learning~\cite{addari2023family,xue2024event,mei2024unsupervised}.
However, standard 3DGS benchmarks assume clear input views, and performance often degrades when images contain transient occlusions such as raindrops on the lens~\cite{liu2025deraings,qian2025weathergs,kulhanek2024wildgaussians}.

To address this issue, several methods ~\cite{li2024derainnerf, qian2025weathergs, liu2025deraings} have been developed to improve 3D reconstruction in raindrop scenes. 
%
%
WeatherGS~\cite{qian2025weathergs} first generates raindrop masks to identify occluded regions and then reconstructs clear scenes by excluding these areas during 3D Gaussian Splatting. 
Meanwhile, DerainGS~\cite{liu2025deraings} incorporates a dedicated image enhancement module to remove raindrop artifacts and employs supervised Gaussian‑ellipsoid fitting, achieving 3D deraining in the final output. 
DerainNeRF~\cite{li2024derainnerf} predicts waterdrop locations with an attention network and trains a Neural Radiance Field to recover the scene. 
RainyScape~\cite{lyu2024rainyscape} couples a predictor network with a learnable latent embedding that captures the scene's rain characteristics. 
Most methods are trained on synthetic raindrops and deliver strong results under the assumption of accurate camera pose estimation and reliable point cloud initialization. However, they overlook the initial disruptions that real raindrops introduce to both pose estimation and point cloud initialization, resulting in poor generalization to real-world raindrop scenarios.

\subsection{3D Raindrop Reconstruction Benchmark and Dataset}
Existing 3D raindrop benchmarks often rely on synthetic blending or limited real-world test data~\cite{li2024derainnerf, liu2025deraings, qian2025weathergs}. Consequently, the locations, shapes, sizes, and quantities of the raindrops remain fixed. Furthermore, these datasets operate under the idealized assumption that cameras can perpetually focus on the background scene, resulting in a significant domain gap from real-world scenarios.
A few real-world test datasets have attempted to simulate rain on camera lenses in stereo or small-scale multi-view setups. 
For example, DerainNeRF~\cite{li2024derainnerf} captures stereo image pairs by spraying water onto a glass plate in front of a calibrated rig, providing binary raindrop masks. WeatherGS~\cite{qian2025weathergs} extracts keyframes from public rainy videos but does not provide rain-free ground truth references. Despite these efforts, existing real-world datasets remain limited in scale and provide insufficient coverage of different focus settings and illumination conditions. To overcome these critical limitations and bridge the domain gap, we introduce OmniRaindrop, the first large-scale, real-world benchmark that provides aligned multi-view data under diverse focal and illumination conditions. 

\begin{figure}[t!]
	\centering
	\includegraphics[width=1.0\linewidth]{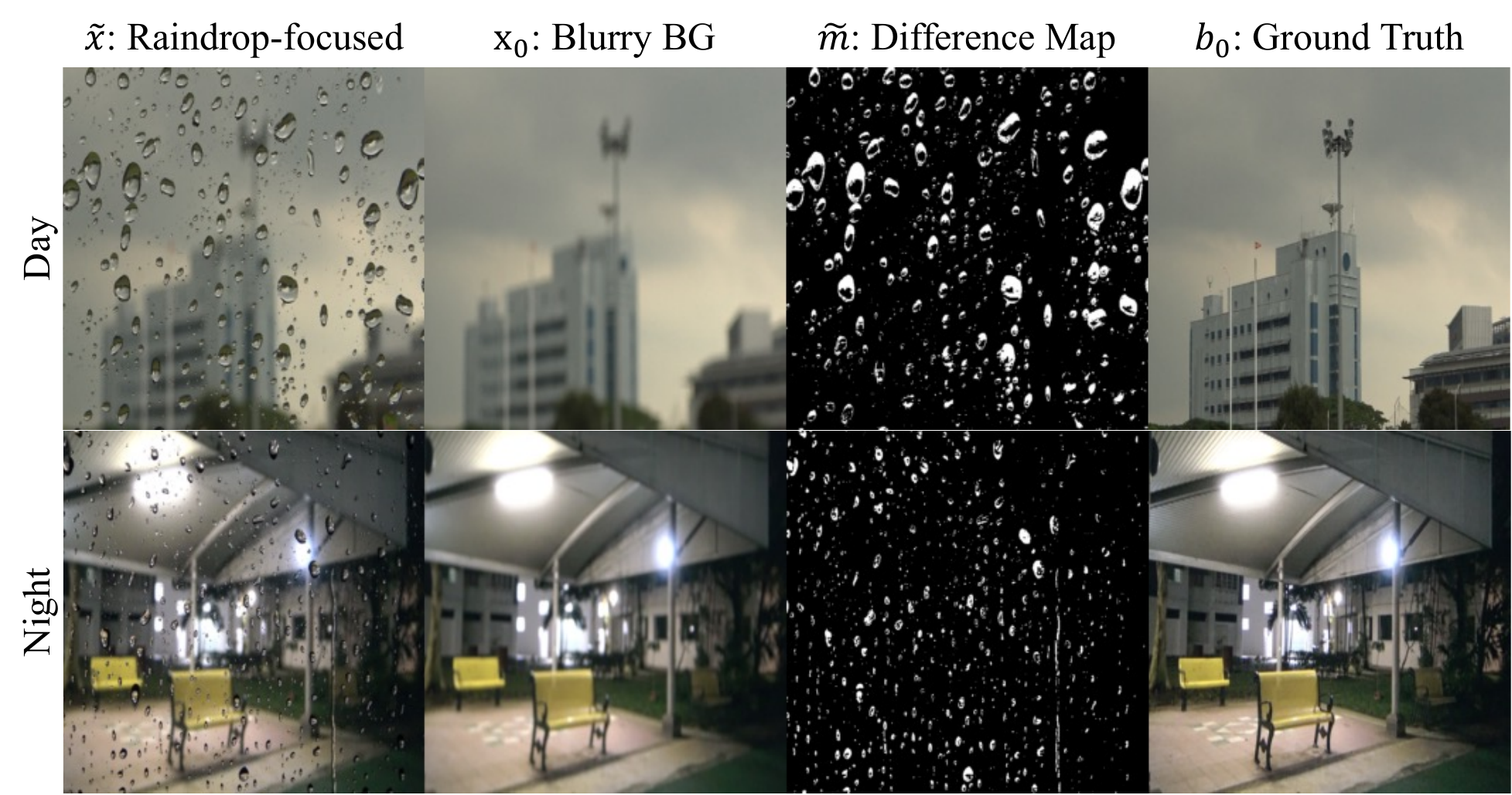}
    \caption{Physical formulation of raindrop-focused degradation. Given a raindrop-focused image $\tilde{x}$ and its corresponding rain-free blurred background $x_0$, the difference map $\tilde{m}$ is computed to isolate raindrop-induced degradation, while $b_0$ denotes the sharp rain-free ground truth.}
	\label{fig:mask}
\end{figure}

\section{OmniRaindrop Dataset and Benchmark}

\begin{figure*}[!ht]  
  \centering
  \includegraphics[width=\textwidth]{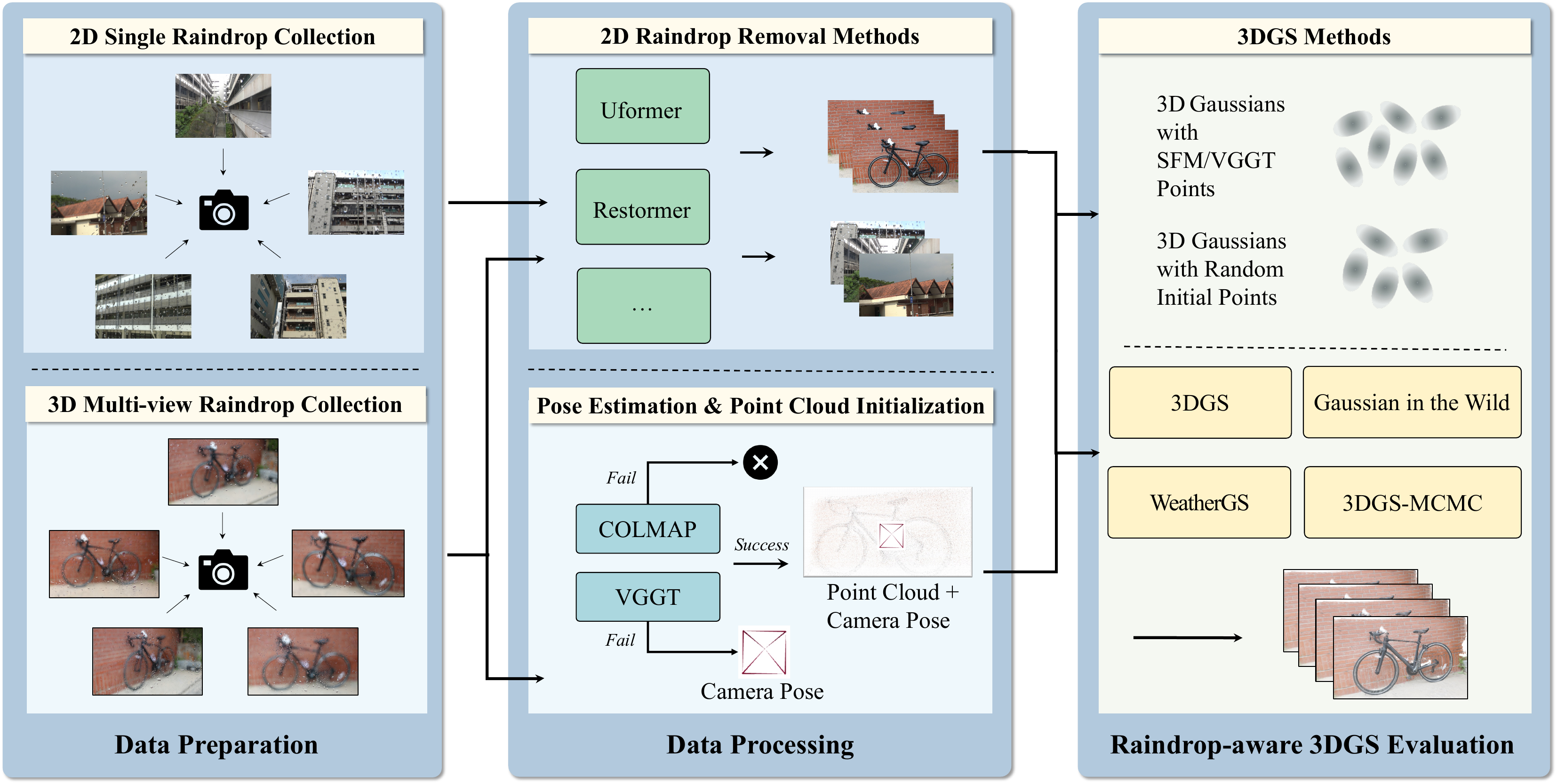}
  \caption{From 2D to 3D raindrop removal benchmark pipeline. We develop the first benchmark for comprehensively evaluating raindrop removal performance. The benchmark begins with dataset collection, proceeds through data processing, and ends with a raindrop-aware 3DGS evaluation. Our benchmark enables the community to explore background-focused and raindrop-focused images, including challenges unique to daytime and nighttime conditions.}
  \label{pipeline}
  \vspace{-3mm}
\end{figure*}

We introduce OmniRaindrop, a large-scale, dual-focused dataset and benchmark for diagnosing the vulnerabilities of visual perception systems under raindrop conditions. As illustrated in Fig.~\ref{pipeline}, it forms a complete pipeline from 2D raindrop-corrupted captures to the final 3D Gaussian reconstruction. In this section, we first describe the physics-constrained acquisition protocol, then report the dataset statistics and properties, and finally present a unified pipeline that assesses both 2D restoration and its cascading impact on 3D reconstruction.

\subsection{Data Acquisition and Physical Formulation}
We first formulate the underlying optical refraction model that governs raindrop interference. Guided by this theoretical foundation, we then design a precise physical acquisition protocol to capture aligned paired and triplet data.

\subsubsection{Optical Model and Focus Shift}
We assume a standard pinhole camera model focused on the background plane~\cite{hartley2003pinhole}. In the absence of optical distortion, all scene elements located on the focal plane appear sharp and well-defined~\cite{hao2019learning}. However, raindrops adhering to a thin transparent cover glass placed directly in front of the lens act as miniature convex lenses. They alter the optical path governed by Snell's Law~\cite{born2013principles}, introducing severe photometric aberrations and local defocus blur~\cite{quan2024deep, jin2024raindrop}.

When the camera is set to be background-focused, we mathematically decompose the raindrop-degraded image $\tilde{\mathbf{x}}$ into a clear background image $\mathbf{b}_0$ following the additive occlusion model~\cite{qian2018attentive}:
\begin{equation}
\tilde{\mathbf{x}} = (1 - \mathbf{M}) \odot \mathbf{x}_0 + \mathbf{D}
\end{equation}
where $\mathbf{D}$ represents the complex illumination refracted by the raindrops from the surrounding environment, and $\odot$ denotes element-wise multiplication. The binary mask $\mathbf{M}(\mathbf{x})=1$ signifies that the pixel belongs to the raindrop-affected region. In this specific focal configuration, the unoccluded background $\mathbf{x}_0$ is structurally identical to the pristine, rain-free ground truth $\mathbf{b}_0$ (i.e., $\mathbf{x}_0 = \mathbf{b}_0$). Consequently, this acquisition yields our aligned background-focused pairs $(\tilde{\mathbf{x}}, \mathbf{b}_0)$.

Conversely, when the camera is set to be raindrop-focused, the image plane captures geometric representations of the raindrop surfaces, while the distant background features suffer from severe depth-of-field blur. In this optical state, the rain-free background $\mathbf{x}_0$ remains substantially blurred, rendering it distinct from the sharp rain-free ground truth ($\mathbf{x}_0 \neq \mathbf{b}_0$). To capture the precise optical distortion without background interference, we compute the pixel-level difference map $\tilde{\mathbf{m}}$ between the raindrop-focused image $\tilde{\mathbf{x}}$ and its corresponding rain-free blurry background $\mathbf{x}_0$:
\begin{equation}
\tilde{\mathbf{m}} = \tilde{\mathbf{x}} - \mathbf{x}_0
\end{equation}
As shown in Fig.~\ref{fig:mask}, this physical decoupling enables the construction of our raindrop-focused triplets $(\tilde{\mathbf{x}}, \mathbf{x}_0, \mathbf{b}_0)$, allowing downstream networks to learn the acquisition-guided degradation decomposition without being misled by dynamic background motion or varying blur kernels.

\subsubsection{Physical Collection Protocol}

To transition the aforementioned theoretical model into a real-world benchmark, scene staticity is important. We employ a heavy-duty pan-tilt sphere platform to keep the camera stationary, minimizing unintended viewpoint changes via fixed tripods. Our optical setup incorporates a pressure sprayer to generate diverse raindrop patterns and a glass plate with high light transmittance mounted ahead of the lens. 

For each static viewpoint, the capture sequence follows the mathematical formulations: We first capture the image with adherent raindrops ($\tilde{\mathbf{x}}$). Subsequently, the glass plate is rapidly removed without perturbing the camera body to acquire the corresponding rain-free background ($\mathbf{x}_0$). Finally, the camera focus is adjusted to infinity to capture the crisp ground-truth background ($\mathbf{b}_0$). This sequential operation is designed to preserve both photometric consistency (e.g., matching day/night illumination) and geometric alignment across the dataset.

\subsection{Dataset Statistics and Properties}
Leveraging the aforementioned physical protocol, we construct OmniRaindrop, a benchmark characterized by large scale, controlled spatial alignment, and diverse illumination conditions.

\textbf{2D Raindrop Subset.} As detailed in Table~\ref{tab:dataset_comparison}, the 2D subset contains 15,186 aligned pairs and triplets, divided into daytime and nighttime domains. The daytime collection has 5,442 samples, with 3,606 raindrop-focused and 1,836 background-focused. The nighttime collection adds 9,744 samples, with 4,838 raindrop-focused and 4,906 background-focused, and poses harder restoration challenges due to severe low-light degradation and complex artificial-light reflections refracted through the raindrops.

\textbf{3D Raindrop Subset.} The 3D subset provides 40 diverse real-world scenes, including 20 daytime scenes and 20 nighttime scenes. Each scene contains at least 28 views, yielding a total of 3,822 high-resolution multi-view images. For every viewpoint within a scene, we capture a photometric triplet, consisting of a raindrop-focused image, a background-focused image, and the corresponding rain-free ground truth. Importantly, the adherent raindrops are physically generated for each capture rather than synthetically composited. As a result, their number, size, shape, and spatial distribution vary naturally across viewpoints, introducing realistic view-dependent occlusion, refraction, and defocus patterns. This property distinguishes OmniRaindrop from prior synthetic 3D raindrop datasets, where raindrop masks are typically rendered or overlaid with fixed or manually controlled patterns. The resulting aligned multi-view triplets provide reliable geometric and photometric references for evaluating raindrop-aware 3DGS reconstruction.

\textbf{Comparison with Existing Datasets.} Compared to existing benchmarks, OmniRaindrop offers broader coverage in both authenticity and scale. Prior 3D raindrop attempts rely on synthetic blending, applying static, rigid raindrop masks to existing clear datasets~\cite{li2024derainnerf, qian2025weathergs, liu2025deraings}. In contrast, OmniRaindrop lets adherent raindrops vary organically in shape, density, and refraction across views, mimicking real-world optical behavior. Compared to previous 2D datasets such as RainDS~\cite{quan2021removing} and Windshield~\cite{soboleva2021raindrops}, it further reduces the synthetic-to-real gap by providing real aligned captures with focal-shift dynamics and nighttime environments.

\begin{table*}[htbp]
  \centering
  \caption{Comprehensive comparison between OmniRaindrop and existing datasets. We compare dataset components, scale, data source, focus condition, and illumination coverage. OmniRaindrop provides a large-scale real-world benchmark with strict alignment for both 2D restoration and 3DGS reconstruction.}
  \label{tab:dataset_comparison}
  \resizebox{1\linewidth}{!}{%
  \begin{tabular}{lcccccccc}
    \toprule
    \multirow{2}{*}{Dataset} & \multirow{2}{*}{Components} & \multicolumn{2}{c}{Scale} & \multirow{2}{*}{Source} & \multicolumn{2}{c}{Bg-focus} & \multicolumn{2}{c}{Rd-focus} \\
    \cmidrule(lr){3-4} \cmidrule(lr){6-7} \cmidrule(lr){8-9}
    & & Scenes & Images & & Day & Night & Day & Night \\
    \midrule
    \multicolumn{9}{l}{\textbf{2D Raindrop Datasets}} \\
    \midrule
    Qian et al.~\cite{qian2018attentive} & 919 Pairs & $-$ & 1,838 & Real & \checkmark & $\times$ & $\times$ & $\times$ \\
    RainDS-Real~\cite{quan2021removing} & 248 Quadruplets & $-$ & 992 & Real & \checkmark & $\times$ & $\times$ & $\times$ \\
    RainDS-Syn~\cite{quan2021removing}  & 1,200 Quadruplets & $-$ & 4,800 & Syn & \checkmark & $\times$ & $\times$ & $\times$ \\
    RobotCar~\cite{porav2019can}  & 4,818 Pairs & $-$ & 9,636 & Real & \checkmark & $\times$ & $\times$ & $\times$ \\
    Windshield~\cite{soboleva2021raindrops}  & Video Frames w/ Masks & $-$ & 3,390 & Real & \checkmark & $\times$ & $\times$ & $\times$ \\
    \midrule
    \textbf{Ours (2D)} & \textbf{8,444 Triplets + 6,742 Pairs} & \textbf{$-$} & \textbf{38,816} & \textbf{Real} & \textbf{\checkmark} & \textbf{\checkmark} & \textbf{\checkmark} & \textbf{\checkmark} \\
    \midrule
    \multicolumn{9}{l}{\textbf{3D Raindrop Datasets}} \\
    \midrule
    DerainNeRF~\cite{li2024derainnerf}  & 20 views / scene & 5  & 100  & Real+Syn & \checkmark & $\times$ & $\times$ & $\times$ \\
    DerainGS(Raindrop Syn)~\cite{liu2025deraings}  & 125--311 views / scene & 7 & 4,878 & Syn & \checkmark & $\times$ & $\times$ & $\times$ \\
    \midrule
    \textbf{Ours (3D)} & \textbf{28--50 views / scene} & \textbf{40} & \textbf{3,822} & \textbf{Real} & \textbf{\checkmark} & \textbf{\checkmark} & \textbf{\checkmark} & \textbf{\checkmark} \\
    \bottomrule
  \end{tabular}%
  }
\end{table*}

\subsection{Unified 2D and 3D Evaluation Pipeline}
To reveal the actual bottlenecks of existing visual perception systems, our benchmark establishes a systematic, end-to-end evaluation pipeline. Instead of merely assessing isolated 2D visual quality, we track how localized photometric distortions propagate into cascading failures within 3DGS reconstruction.

\subsubsection{2D Raindrop Removal Baselines}
We evaluate single-image raindrop removal through a dual-track evaluation paradigm. First, these models are assessed independently on the pure 2D image restoration task. Second, since traditional 3DGS methods lack inherent raindrop-handling capabilities, these exact 2D baselines simultaneously serve as the critical pre-processing front-end for our 3D reconstruction pipeline.

We select a suite of restoration models that balance domain-specific designs with architectural advances. For specialized deraining, we use AtGAN~\cite{qian2018attentive}, RainDS~\cite{quan2021removing}, and IDT~\cite{xiao2022IDT}. To test whether modern generative modeling can overcome structural background loss, we add state-of-the-art backbones with long-range dependency modeling: Restormer~\cite{zamir2022restormer}, Uformer~\cite{wang2022uformer}, RDiffusion~\cite{ozdenizci2023restoring}, and the diffusion-based DiT~\cite{peebles2023scalable}. For a fair comparison, all baselines are trained on our 2D subset~\cite{jin2024raindrop}. This setup first quantifies their standalone 2D recovery, then evaluates how their outputs affect the downstream 3DGS reconstruction.

\subsubsection{3D Pose Estimation and Point Cloud Initialization}
A common simplifying assumption in existing 3DGS raindrop research is the availability of accurate camera poses and dense initial point clouds. Our benchmark deliberately exposes the raw 3D reconstruction pipeline to severe raindrop interference. We apply COLMAP~\cite{schonberger2016structure}—leveraging Structure-from-Motion (SfM)~\cite{schonberger2016structure} for intrinsic/extrinsic parameters and Multi-View Stereo (MVS)~\cite{furukawa2015MVS} for initial point clouds—and Visual Geometry Grounded Transformer (VGGT)~\cite{wang2025vggt}, which utilizes robust DINO-based features~\cite{oquab2023dinov2}, to the unconstrained multi-view images. Furthermore, for extreme raindrop-focused scenes where severe optical blur causes complete initialization collapse (i.e., yielding zero matchable points), we implement a random point cloud initialization strategy generating initial points to assess the extreme convergence capabilities of downstream 3DGS optimizers.

\subsubsection{Raindrop-aware 3DGS Evaluation}
Utilizing the estimated poses and pre-processed multi-view images (enhanced by 2D restoration front-ends such as Uformer~\cite{wang2022uformer} or Restormer~\cite{zamir2022restormer}), we evaluate four representative 3DGS architectures. These include standard 3DGS~\cite{kerbl20233d} as the fundamental baseline, WeatherGS~\cite{qian2025weathergs} which incorporates dedicated occlusion masks, GS-W~\cite{zhang2024GS-W} designed specifically for unconstrained and view-inconsistent image collections, and 3DGS-MCMC~\cite{kheradmand2024MCMCGS} which demonstrates theoretical robustness to sparse or noisy point cloud initializations. This rigorous framework allows us to evaluate how early-stage photometric artifacts dictate the geometric fidelity of the final 3D scene representation.

\begin{table*}[!t]
\centering
\small
\setlength{\tabcolsep}{5pt}
\renewcommand{\arraystretch}{1.15}

\caption{Quantitative evaluation of representative raindrop removal methods
and general image restoration backbones on the daytime and nighttime subsets
of OmniRaindrop. We report PSNR, SSIM, and LPIPS to evaluate pixel fidelity,
structural similarity, and perceptual quality, respectively. The best results
are highlighted for each metric.}
\label{tab:PSNR_SSIM}

\begin{tabular}{@{}llcccccc|ccc@{}}
\toprule
\multicolumn{2}{c}{\multirow{2}{*}{Data/Metric}}
&
\multirow{2}{*}{Input}
&
\multicolumn{5}{c|}{Raindrop Algorithms}
&
\multicolumn{3}{c}{Restoration Backbones}
\\
\cmidrule(lr){4-8}
\cmidrule(l){9-11}

\multicolumn{2}{c}{}
&
&
AtGAN
&
Robot
&
RainDS
&
IDT
&
RDiff
&
Restormer
&
Uformer
&
DiT
\\
\midrule

\multirow{3}{*}{Day}
& PSNR$\uparrow$
& 21.92 & 23.62 & 23.82 & 25.39 & 26.05 & 25.52
& \textbf{26.08} & 25.78 & 26.03
\\
\cmidrule(l){2-11}

& SSIM$\uparrow$
& 0.560 & 0.658 & 0.670 & 0.717 & 0.736 & 0.734
& 0.748 & 0.727 & \textbf{0.752}
\\
\cmidrule(l){2-11}

& LPIPS$\downarrow$
& 0.247 & 0.200 & 0.191 & 0.147 & 0.141 & 0.111
& 0.131 & 0.148 & \textbf{0.106}
\\
\midrule

\multirow{3}{*}{Night}
& PSNR$\uparrow$
& 24.78 & 24.38 & 24.57 & 25.39 & 26.81 & 26.48
& \textbf{26.87} & 25.26 & 26.23
\\
\cmidrule(l){2-11}

& SSIM$\uparrow$
& 0.726 & 0.773 & 0.787 & 0.820 & \textbf{0.851} & 0.831
& 0.848 & 0.816 & 0.826
\\
\cmidrule(l){2-11}

& LPIPS$\downarrow$
& 0.209 & 0.185 & 0.182 & 0.139 & 0.125 & 0.112
& 0.123 & 0.142 & \textbf{0.111}
\\
\bottomrule
\end{tabular}
\end{table*}

\begin{figure*}[t!]
	\centering
	\includegraphics[width=1.0\linewidth]{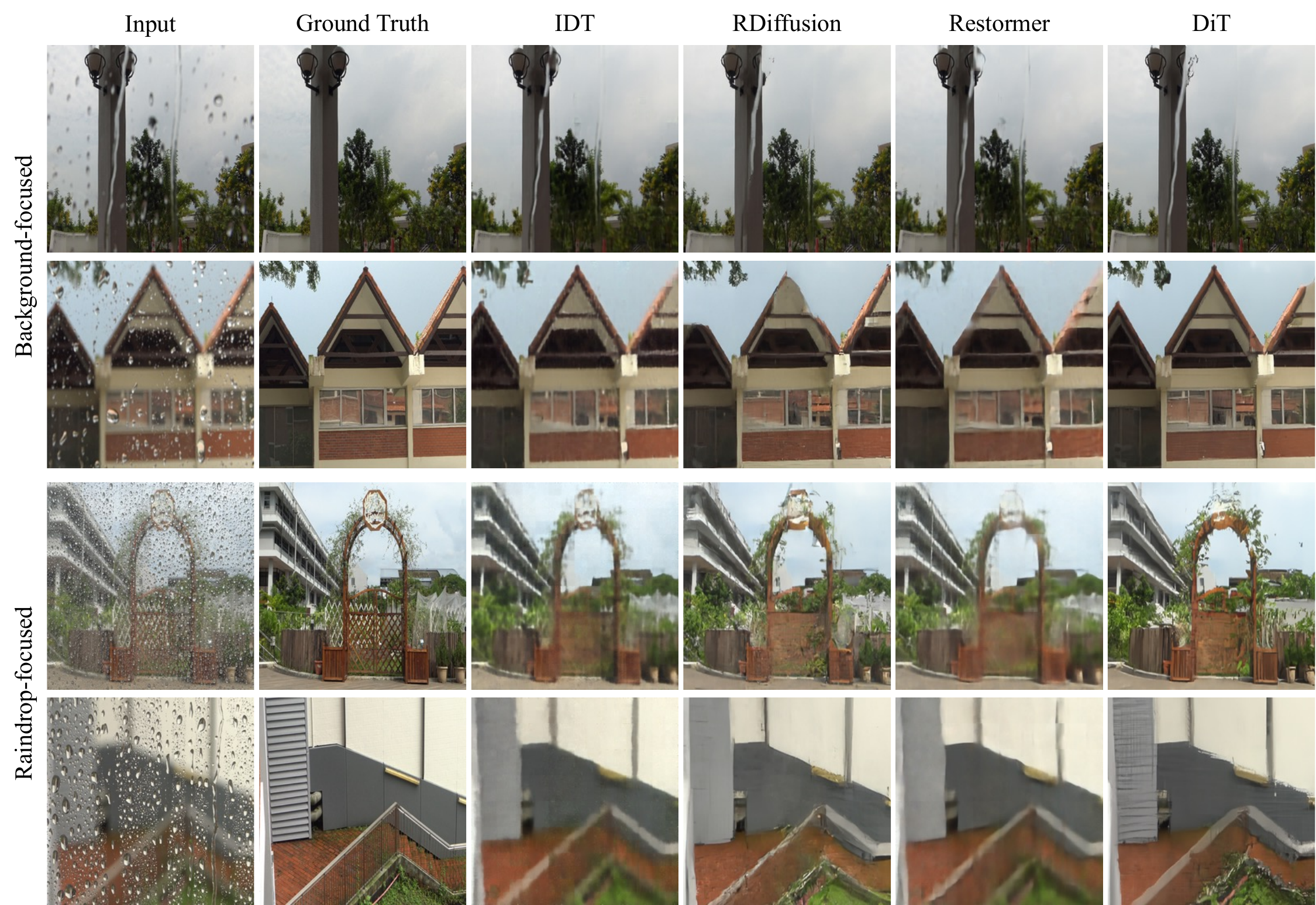}	
	\caption{Qualitative comparison of representative raindrop removal methods on daytime images. Existing methods remove some diffuse occlusions in background-focused inputs, but struggle with the sharper raindrops and severe background blur in raindrop-focused inputs.}
	\label{fig:dayrd}
\end{figure*}

\section{Experiments}

\subsection{Implementation Details}

For 2D raindrop removal comparison, we conduct experiments using the original implementation of these methods. All experiments are
implemented in PyTorch and are trained on four NVIDIA RTX A5000 GPUs. For 3DGS data processing, the original scene images are captured at a resolution of $1024 \times 576$, while all 3DGS methods are reconstructed using images downsampled to one-quarter resolution, i.e., $256 \times 144$, for computational efficiency and fair comparison. During VGGT-based camera pose estimation and point cloud initialization, we follow the preprocessing protocol of VGGT~\cite{wang2025vggt}, resizing each input image to $518 \times 518$ before processing. Likewise, for Restormer~\cite{zamir2022restormer} and IDT~\cite{xiao2022IDT}, we downscale images to $128 \times 128$; for Uformer~\cite{wang2022uformer} and AtGAN~\cite{qian2018attentive}, we downscale images to $256 \times 256$, and then tile the results to restore the original resolution.

For all 3DGS methods, we follow the official optimization settings: 30,000 iterations for 3DGS, WeatherGS, and 3DGS-MCMC, and 70,000 iterations for GS-W. For 3DGS-MCMC, we set the initial point cloud size to 100,000, based on the number of points that VGGT can initialize in our dataset. All models are implemented in PyTorch and evaluated on eight NVIDIA RTX 3090 GPUs.

\subsection{2D Raindrop Removal Evaluation}
\label{2D raindrop removal}
\noindent\textbf{Quantitative Comparison.}

\begin{figure*}[t!]
	\centering
	\includegraphics[width=1.0\linewidth]{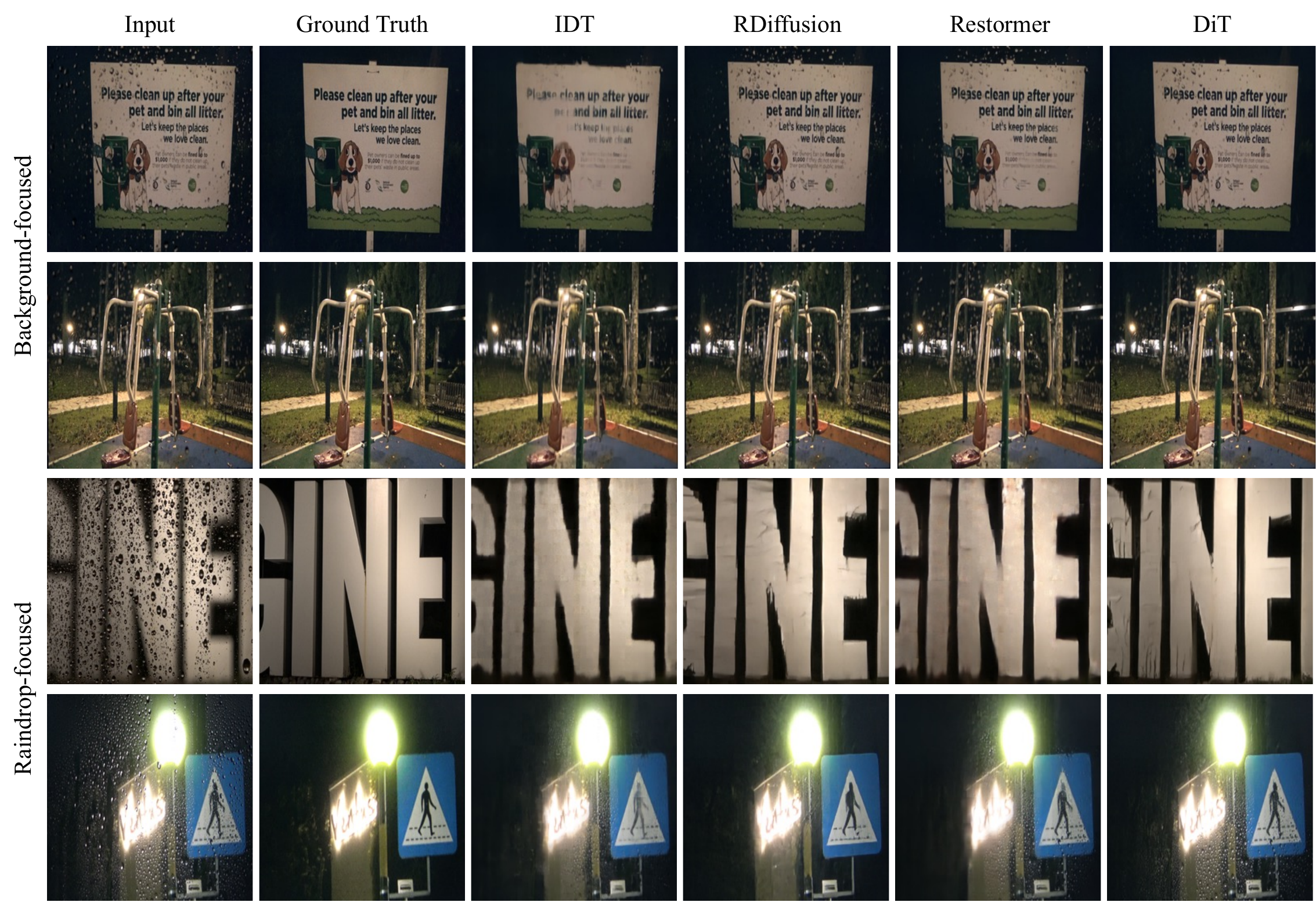}
	\caption{Qualitative comparison of representative raindrop removal methods on nighttime images. Compared with daytime scenes, nighttime raindrops introduce stronger light refraction, specular highlights, halos, local saturation, and non-linear intensity variations. Existing methods often leave residual raindrops or over-smooth background details in background-focused images. Under raindrop-focused inputs, the combined effect of sharp raindrop structures, severe background defocus, and nighttime illumination distortion makes it more difficult to recover clean and structurally faithful backgrounds.}
	\label{fig:nightrd}
\end{figure*} 

We use three popular metrics PSNR, SSIM~\cite{wang2004image}, and LPIPS~\cite{zhang2018unreasonable} in generation tasks~\cite{jin2021dc,jin2022unsupervised,jin2023enhancing,jin2024des3,ye2022perceiving,chen2023sparse,chen2023msp,ye2023adverse,chen2022snowformer,wang2024restoring,wang2023seeing,li2024sed,li2023learning,li2025diffusion,li2020learning,luo2023crefnet,luo2024intrinsicdiffusion,wang2024correlation,wang2023promptrestorer}. 
The quantitative results are shown in Table \ref{tab:PSNR_SSIM}. 
For the daytime dataset, the PSNR and SSIM of the original input images are 21.92 and 0.560. 
Raindrop algorithm IDT~\cite{xiao2022IDT} achieves the best performance among raindrop-specific methods, with a PSNR of 26.05 dB and an SSIM of 0.736.
For the restoration backbones, DiT~\cite{peebles2023scalable} achieves a PSNR of 26.03 and a SSIM of 0.752.
Restormer~\cite{zamir2022restormer} achieves a PSNR of 26.08 and a SSIM of 0.748, which also performs competitively.

It can be observed that, for the nighttime dataset, PSNR and SSIM of the input images are 24.78 and 0.726. 
Among the raindrop algorithms, IDT~\cite{xiao2022IDT} achieves better performance than other methods, with a PSNR of 26.81 and a SSIM of 0.851.
For the restoration backbones,  Restormer~\cite{zamir2022restormer} achieves the highest PSNR of 26.87, while DiT~\cite{peebles2023scalable} achieves the lowest LPIPS of 0.111, which outperforms other backbones.
 
Overall, general-purpose restoration backbones perform comparably to or better than specialized raindrop-removal methods, although no single method consistently dominates across all metrics and illumination conditions. 

\noindent\textbf{Qualitative Comparison.}
Figs.~\ref{fig:dayrd}-\ref{fig:nightrd} show the qualitative results of state-of-the-art methods. 
In Fig.~\ref{fig:dayrd}, for daytime background-focused inputs, we can observe that existing state-of-the-art methods remove many visible raindrops. However, elongated raindrop traces are often only partially detected and therefore remain in the restored images. 
Furthermore, in Fig.~\ref{fig:nightrd}, we find that existing state-of-the-art methods struggle to handle nighttime background-focused samples. To be specific, residual raindrops remain in these samples, and the restored image details are unclear, since nighttime raindrops usually reflect artificial light and thus are more complex than daytime raindrops. Since many existing methods are developed and evaluated primarily on daytime data, their ability to handle complex nighttime illumination distortions remains limited.

\noindent\textbf{Analysis of 2D Restoration Difficulty.}
We evaluate both raindrop-specific methods and general-purpose restoration backbones on the daytime and nighttime subsets. Restoration backbones such as Restormer~\cite{zamir2022restormer} and DiT~\cite{peebles2023scalable} achieve competitive performance, showing the benefit of stronger global modeling and restoration capacity. Among raindrop-specific methods, IDT~\cite{xiao2022IDT} performs strongly, suggesting that degradation-aware designs remain useful for adherent raindrop removal.
The major difficulty of OmniRaindrop comes from focus-dependent degradation. In background-focused images, the background remains relatively sharp, while raindrops appear as diffuse, out-of-focus occlusions. In contrast, raindrop-focused images contain sharper raindrop structures but suffer from severe background defocus.
Therefore, raindrop-focused restoration is not merely a raindrop removal problem. It requires simultaneous removal of sharp occlusions, recovery of defocused background structures, and reconstruction of missing high-frequency textures.
Prior studies that explicitly model defocus also show that regions with different defocus levels exhibit different texture statistics, supporting our evaluation design across different focus settings~\cite{xing2023daqe}.
Nighttime scenes introduce a distinct illumination-dependent degradation regime. Artificial light sources are refracted by adherent raindrops, producing highlights, halos, local saturation, and non-linear intensity changes. These effects are fundamentally different from daytime occlusion and blur.
As a result, nighttime raindrop removal requires models to handle not only geometric occlusion and defocus blur, but also complex illumination distortions caused by light refraction.

\begin{table*}[t]
  \centering
  \small
  \renewcommand{\arraystretch}{1.2}
  
  \caption{Quantitative comparison of four 3DGS methods on the 3D subset using VGGT~\cite{wang2025vggt} for camera pose estimation and point cloud initialization. For each daytime/nighttime split, results are averaged over all background-focused and raindrop-focused cases. Uformer~\cite{wang2022uformer}, Restormer~\cite{zamir2022restormer}, and IDT~\cite{xiao2022IDT} denote different 2D restoration front-ends used before 3DGS reconstruction, while WeatherGS~\cite{qian2025weathergs} is evaluated with its original pipeline.}
  \label{tab:quan-baseline-vggt-day-night}

  \resizebox{\textwidth}{!}{
  \begin{tabular}{@{}
      l l
      c c c
      c c c
      c c c
      c
    @{}}
    \toprule
      &
      & \multicolumn{3}{c}{3DGS}
      & \multicolumn{3}{c}{3DGS-MCMC}
      & \multicolumn{3}{c}{GS-W}
      & WeatherGS \\
    \cmidrule(lr){3-5} \cmidrule(lr){6-8} \cmidrule(lr){9-11} \cmidrule(lr){12-12}
    Scene & Metrics
      & Uformer & Restormer & IDT
      & Uformer & Restormer & IDT
      & Uformer & Restormer & IDT
      & / \\
    \midrule

    \multirow{3}{*}{Day}
      & PSNR $\uparrow$
      & 17.119 & 17.162 & 17.211
      & 18.202 & 18.217 & 18.217
      & 19.005 & 18.983 & \textbf{19.192}
      & 16.918 \\
      & SSIM $\uparrow$
      & 0.468 & 0.470 & 0.476
      & 0.501 & 0.494 & 0.502
      & 0.540 & 0.538 & \textbf{0.550}
      & 0.443 \\
      & LPIPS $\downarrow$
      & 0.442 & 0.439 & 0.434
      & 0.419 & 0.419 & 0.414
      & 0.355 & 0.355 & \textbf{0.340}
      & 0.446 \\

    \midrule

    \multirow{3}{*}{Night}
      & PSNR $\uparrow$
      & 20.633 & 20.744 & 20.862
      & 21.222 & 21.476 & 21.537
      & 22.254 & \textbf{22.268} & 22.162
      & 20.672 \\
      & SSIM $\uparrow$
      & 0.536 & 0.540 & 0.544
      & \textbf{0.622} & 0.620 & \textbf{0.622}
      & 0.596 & 0.588 & 0.580
      & 0.518 \\
      & LPIPS $\downarrow$
      & 0.401 & 0.397 & 0.396
      & 0.386 & 0.380 & \textbf{0.368}
      & 0.371 & 0.372 & 0.374
      & 0.405 \\

    \bottomrule
  \end{tabular}
  }
  \vspace{-3mm}
\end{table*}

Raindrop-specific methods are usually designed to detect and remove localized adherent raindrop regions, and thus can be effective when the background is relatively sharp. However, they often rely on the assumption that the underlying background structure is observable, which becomes invalid in raindrop-focused images with severe defocus blur.
In contrast, general restoration backbones provide stronger deblurring and texture recovery capability, but they do not explicitly model raindrop refraction, occlusion geometry, or focus-dependent degradation. Consequently, they may either leave residual raindrops or hallucinate inconsistent textures in heavily degraded regions.

These observations indicate that single-image restoration quality alone is insufficient for evaluating raindrop removal in 3D reconstruction scenarios. Even when a restoration model improves PSNR or perceptual quality on individual views, it may introduce view-dependent textures, inconsistent edges, or hallucinated details across different viewpoints. Such inconsistencies can further affect feature matching, pose estimation, point cloud initialization, and 3DGS optimization.
Therefore, we further conduct a pipeline-level analysis to examine how 2D restoration outputs propagate into camera pose estimation, point cloud initialization, and final 3DGS reconstruction.

\subsection{3DGS Reconstruction Analysis}

\noindent\textbf{Quantitative Comparison.}
Table~\ref{tab:quan-baseline-vggt-day-night} reports the quantitative reconstruction results of representative 3DGS methods on the 3D subset. We evaluate standard 3DGS~\cite{kerbl20233d}, 3DGS-MCMC~\cite{kheradmand2024MCMCGS}, GS-W~\cite{zhang2024GS-W}, and WeatherGS~\cite{qian2025weathergs} using PSNR, SSIM, and LPIPS, where higher PSNR/SSIM and lower LPIPS indicate better rendering quality. For 3DGS, 3DGS-MCMC, and GS-W, we further compare three 2D restoration methods, including Uformer~\cite{wang2022uformer}, Restormer~\cite{zamir2022restormer}, and IDT~\cite{xiao2022IDT}. All methods use VGGT-based camera poses and point cloud initialization for a fair comparison.

On daytime scenes, GS-W achieves the best overall performance. In particular, GS-W with IDT obtains the highest PSNR of 19.192 dB, the highest SSIM of 0.550, and the lowest LPIPS of 0.340. 3DGS-MCMC generally improves over standard 3DGS across different restoration methods, while WeatherGS performs below the strongest GS-W and 3DGS-MCMC. These results indicate that methods designed to handle unconstrained or unstable multi-view inputs are more suitable for raindrop-contaminated reconstruction.

On nighttime scenes, GS-W remains competitive and achieves the highest PSNR with Restormer, reaching 22.268 dB, while the lowest LPIPS is obtained by 3DGS-MCMC with IDT, reaching 0.368. The highest SSIM is achieved by 3DGS-MCMC, with Uformer and IDT both reaching 0.622 after rounding. This metric-dependent behavior suggests that nighttime reconstruction involves a trade-off between perceptual quality, pixel-level fidelity, and structural consistency.

Overall, GS-W shows strong reconstruction performance, especially in terms of PSNR, while 3DGS-MCMC provides competitive structural and perceptual performance in nighttime settings. The choice of 2D restoration method also affects final reconstruction quality, but the best 2D restoration model does not always lead to the best 3D reconstruction result. We analyze this discrepancy in the following section.

\begin{figure*}[t]
      \centering
      \includegraphics[width=1.0\linewidth]{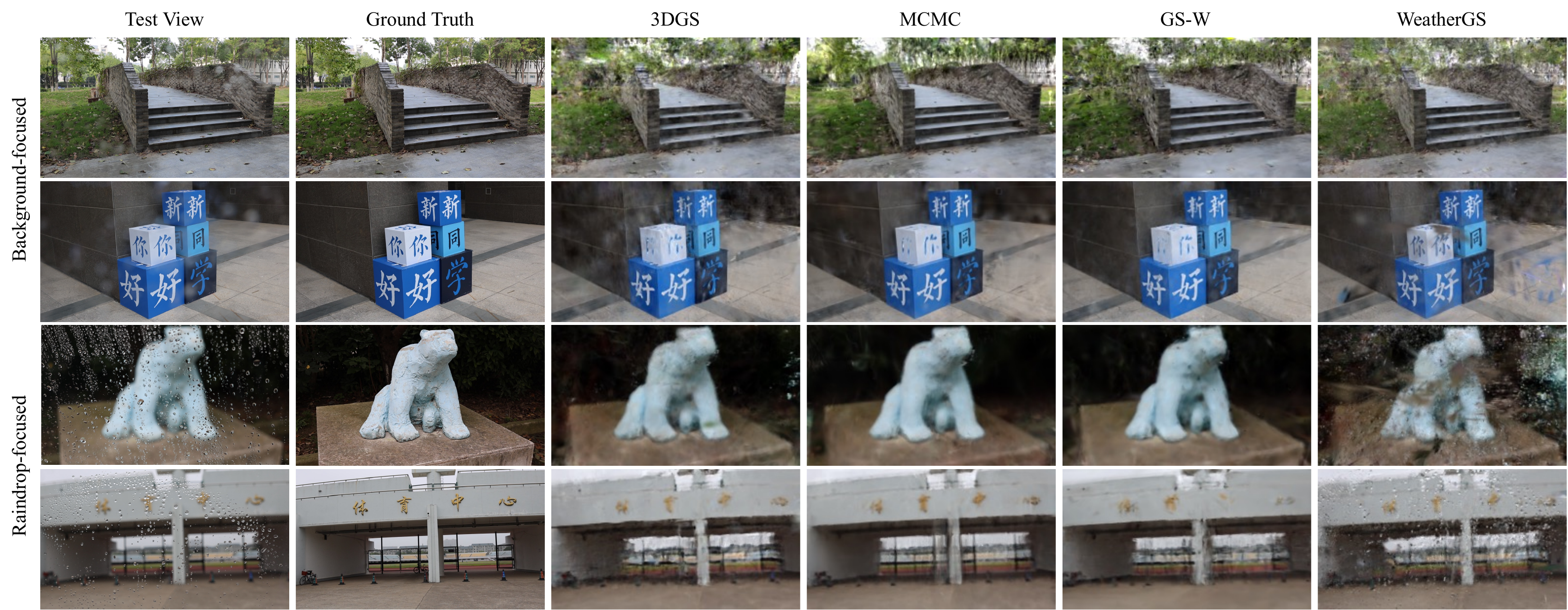}
      \caption{Qualitative comparison of four 3DGS methods under raindrop-corrupted inputs. Background-focused inputs generally preserve more scene details and lead to higher reconstruction quality than raindrop-focused inputs. WeatherGS~\cite{qian2025weathergs} produces relatively smooth reconstructions for blurred inputs, while GS-W~\cite{zhang2024GS-W} produces the most visually coherent results in these examples.}
      \label{3Dday}
\end{figure*}

\begin{figure*}[t]
      \centering
      \includegraphics[width=1.0\linewidth]{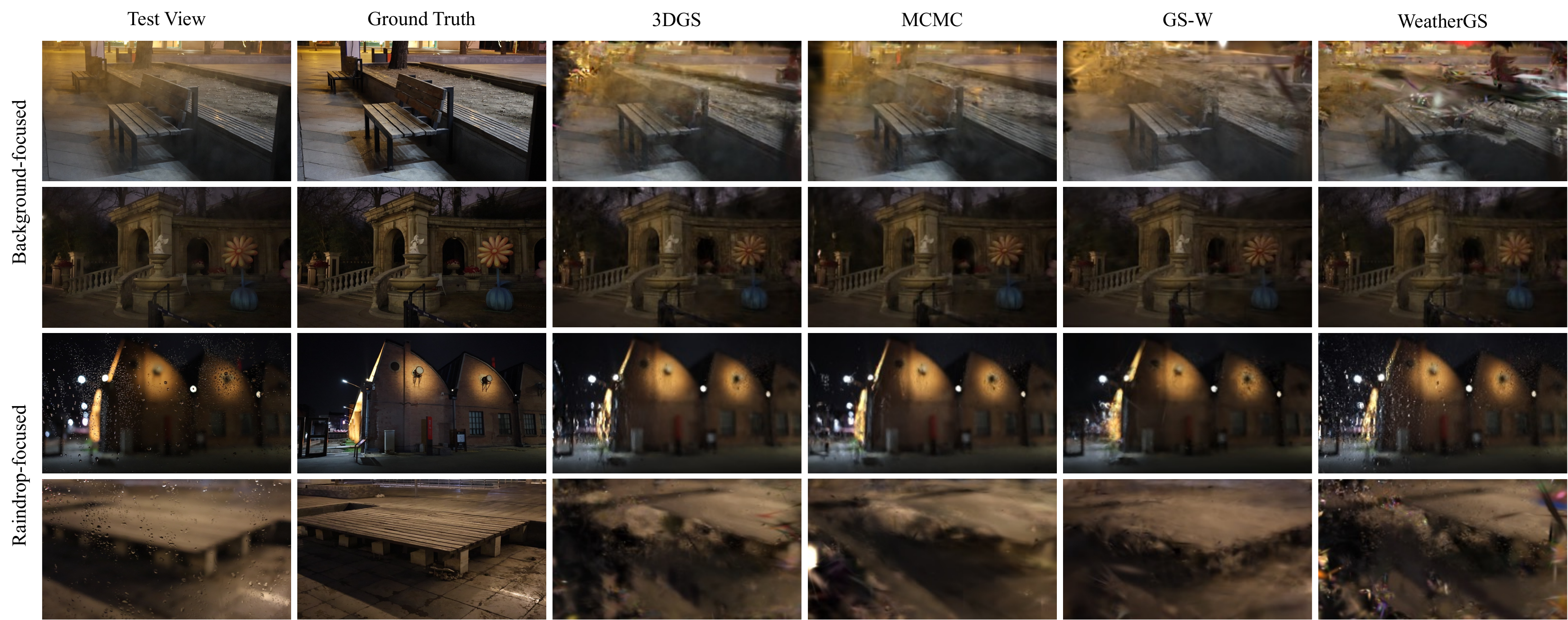}
      \caption{Qualitative comparison of four 3DGS methods under nighttime raindrop-corrupted inputs. Nighttime scenes introduce complex artificial-light refraction, highlights, and low-light degradation, which can introduce inconsistencies across views. Compared with background-focused inputs, raindrop-focused inputs lead to more severe structural degradation due to sharp raindrop interference and defocused background observations.}
      \label{3Dnight}
\end{figure*}

\noindent\textbf{Qualitative Comparison.}
Fig.~\ref{3Dday} shows the qualitative comparison on daytime scenes. Background-focused inputs generally preserve clearer scene structures and lead to sharper reconstruction results. In contrast, raindrop-focused inputs produce more visible artifacts, including blurred geometry, residual raindrop-like textures, floaters, and incomplete structures. Among the compared methods, GS-W~\cite{zhang2024GS-W} produces the most visually coherent results, while 3DGS-MCMC~\cite{kheradmand2024MCMCGS} is more stable than standard 3DGS. WeatherGS~\cite{qian2025weathergs} tends to generate smoother reconstructions but often loses fine structural details.

Fig.~\ref{3Dnight} presents the qualitative results on nighttime scenes. Raindrop-induced refraction of artificial light introduces highlights, halos, local saturation, and low-light artifacts, which can produce viewdependent errors in the reconstructed scenes. These effects lead to stronger view-dependent artifacts in the reconstructed scenes. The degradation is especially severe for raindrop-focused inputs, where sharp raindrop structures and defocused backgrounds jointly cause distorted geometry and incomplete appearance recovery. Although GS-W remains visually more robust, all methods still suffer from noticeable quality degradation under nighttime raindrop interference.

The qualitative results in Fig.~\ref{3Dday} and Fig.~\ref{3Dnight} show that background-focused inputs generally preserve clearer scene structures and lead to fewer reconstruction artifacts than raindrop-focused inputs. Nighttime scenes additionally exhibit illumination-dependent reconstruction artifacts, as artificial-light refraction, highlights, and low-light degradation increase view-dependent inconsistencies. 
Together, these qualitative observations and the results in Table ~\ref{tab:quan-baseline-vggt-day-night} show that current 3DGS methods remain sensitive to real adherent-raindrop degradation across different illumination conditions.

\subsection{Analysis of 3DGS Reconstruction Degradation}

\noindent\textbf{Single-view Restoration Does Not Guarantee Better 3D Reconstruction.}
2D restoration affects final 3DGS quality, but its influence is not equivalent to its single-image restoration performance. Existing raindrop removal methods restore each input view independently, so they can improve per-view fidelity while still producing view-dependent residual raindrops, inconsistent boundaries, or hallucinated textures across different views. These artifacts are less visible in single-image evaluation but are harmful to 3DGS, which optimizes a shared scene representation from multi-view photometric consistency. As a result, a method that improves PSNR, SSIM, or LPIPS on individual images may still introduce inconsistent evidence for Gaussian optimization, leading to floaters, blurred geometry, and unstable appearance in the reconstructed scene. This explains why the best 2D restoration model does not necessarily yield the best 3D reconstruction result.

\begin{table}[t]
\centering
\caption{Comparison of point cloud initialization generated by COLMAP~\cite{schonberger2016structure} and VGGT~\cite{wang2025vggt} under background-focus and raindrop-focus inputs. Empty scenes denote completed runs that produce zero point or invalid initial point cloud files. Average point counts are computed only over scenes with valid point clouds. }
\label{tab:point_cloud_comparison}
\begin{tabular}{crrrr}
\toprule
\multirowcell{3}{Method} & \multicolumn{2}{c}{background-} & \multicolumn{2}{c}{raindrop-} \\
 & \multicolumn{2}{c}{focus} & \multicolumn{2}{c}{focus} \\
\cmidrule(lr){2-3} \cmidrule(lr){4-5}
 & \makecell{Empty\\Scene} & \makecell{Avg.\\Points} & \makecell{Empty\\Scene} & \makecell{Avg.\\Points} \\
\midrule
COLMAP & 0/40 & 6,199.55 & 1/40 & 1,634.54 \\
VGGT   & 9/40 & 82,069.97 & 37/40 & 6,402.00 \\
\bottomrule
\end{tabular}
\end{table}

\begin{figure*}[t!]
    \centering
    \includegraphics[width=1\linewidth]{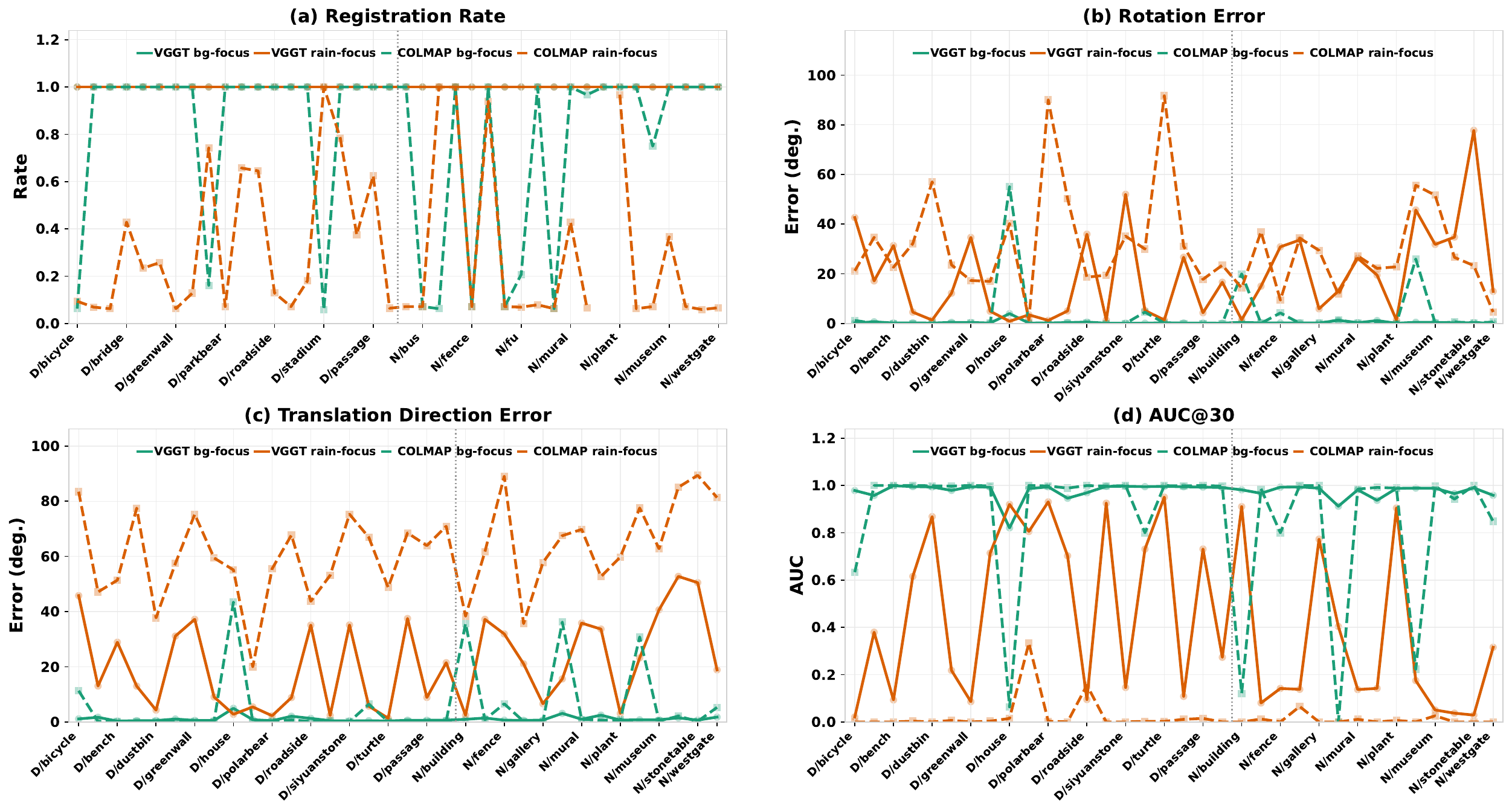}
    \caption{Camera pose quality comparison between COLMAP~\cite{schonberger2016structure} and VGGT~\cite{wang2025vggt} under background-focused and raindrop-focused inputs. We report four metrics across representative daytime and nighttime scenes: (a) registration rate, (b) rotation error, (c) translation error, and (d) AUC@30. Raindrop-focused inputs substantially reduce registration stability and pose accuracy, especially for COLMAP, while VGGT shows stronger robustness but still degrades under severe raindrop-focused conditions.}    
    \label{fig:camera_pose_quality}
\end{figure*}

\noindent\textbf{Raindrops Corrupt Geometric Initialization.}
The degradation also appears before 3DGS optimization, at the stage of camera pose estimation and point cloud initialization. As shown in Fig.~\ref{fig:camera_pose_quality}, raindrop-focused inputs reduce registration stability and pose accuracy, especially for COLMAP~\cite{schonberger2016structure}. This trend is consistent with the point cloud statistics in Table~\ref{tab:point_cloud_comparison}: COLMAP completes almost all scenes, but its average point count drops sharply from the background focus setting to the raindrop focus. VGGT~\cite{wang2025vggt} completes the full camera pose estimation and point cloud initialization pipeline for all scenes under both focus settings, but some completed runs produce zero point. Among cases with valid point clouds, VGGT generates much denser initial point clouds than COLMAP. The pose quality curves further show that a dense initialization alone does not guarantee accurate geometry.


This behavior may reflect the different optical constraints of the two focus settings. In background-focused images, the scene background remains relatively observable, while raindrops mainly appear as diffuse local occlusions. Many stable scene structures therefore remain available for cross-view matching. In raindrop-focused images, however, the sharp raindrop patterns are view-dependent artifacts attached to the imaging surface, whereas the actual scene background becomes severely defocused. These observations provide few reliable correspondences for fixed 3D scene points, increasing pose ambiguity and weakening triangulation. Thus, adherent raindrops should not be treated only as photometric corruption: they also introduce geometric degradation that affects Gaussian placement, densification, and rendering quality.

\noindent\textbf{Method Limitations under Raindrop-induced Degradation.}
The different behaviors of the evaluated 3DGS methods suggest that each method addresses only part of the degradation chain. Standard 3DGS assumes accurate poses, reliable initialization, and multi-view consistent observations; all three assumptions are weakened under real adherent raindrops. 3DGS-MCMC~\cite{kheradmand2024MCMCGS} improves robustness to initialization through sampling-based Gaussian optimization, which explains its stronger structural stability in several settings, but it does not explicitly model raindrop refraction, focus-dependent blur, or cross-view restoration inconsistency. WeatherGS~\cite{qian2025weathergs} introduces weather-aware modeling and can suppress part of the corrupted appearance, yet adherent raindrops are more complex than mask-like weather occlusions because they combine refraction, local illumination distortion, and defocus blur. GS-W~\cite{zhang2024GS-W} is more suitable for view-inconsistent inputs and therefore achieves the strongest PSNR-oriented performance and visually coherent results in several settings, but it still does not explicitly separate stable scene content from raindrop-induced artifacts.

Together, these observations indicate that raindrop-aware 3D reconstruction requires a joint treatment of image restoration, geometric initialization, and 3DGS optimization. Improving only one stage can alleviate part of the degradation, but it may leave other failure sources unresolved. To further disentangle these coupled effects, we next conduct an ablation study that independently varies the input appearance and the source of geometric initialization.

\subsection{Ablation Study}

\noindent\textbf{Experimental Design.}
The analysis above shows that raindrop degradation affects both image observations and geometric initialization. To disentangle these factors, we vary two axes: the input images and the source of geometric initialization. As shown in Table~\ref{tab:ablation_appearance_geometry}, the default setting uses IDT-restored~\cite{xiao2022IDT} images as the appearance inputs for 3DGS optimization, while the camera poses and initial point clouds are estimated from the corresponding original, unprocessed raindrop-corrupted images. The geometry oracle keeps the IDT-restored appearance inputs but replaces the raindrop-image-derived geometry with geometry estimated from the rain-free GT images, isolating the effect of geometric initialization errors. The appearance oracle replaces the inputs with rain-free GT images while keeping raindrop-image geometry, isolating the effect of appearance degradation. The full oracle uses both GT images and GT-image geometry, providing an empirical upper bound under the same reconstruction pipeline. Here, GT-image geometry is not sensor-level ground-truth geometry; it denotes poses and point clouds estimated from rain-free GT images using the same initialization pipeline.

\begin{table}[t]
\centering
\small
\renewcommand{\arraystretch}{1.15}
\setlength{\tabcolsep}{3pt}

\caption{Ablation study of appearance inputs and geometry initialization for 3DGS reconstruction. In the Input Image column, “Raindrop” denotes raindrop-corrupted images restored by IDT~\cite{xiao2022IDT} before being used for 3DGS optimization, whereas “GT” denotes the corresponding rain-free ground-truth images. In the Geometry Initialization column, “Raindrop” and “GT” indicate that camera poses and initial point clouds are estimated from the original, unprocessed raindrop-corrupted images and the rain-free ground-truth images, respectively.}
\label{tab:ablation_appearance_geometry}

\begin{tabular}{@{}llccc@{}}
\toprule
\makecell{Input\\Image}
&
\makecell{Geometry\\Initialization}
&
PSNR$\uparrow$
&
SSIM$\uparrow$
&
LPIPS$\downarrow$
\\
\midrule
Raindrop & Raindrop & 19.037 & 0.510 & 0.415 \\
Raindrop & GT-image & 21.310 & 0.620 & 0.363 \\
GT       & Raindrop & 20.464 & 0.551 & 0.333 \\
GT       & GT-image & 25.525 & 0.811 & 0.156 \\
\bottomrule
\end{tabular}

\end{table}

\noindent\textbf{Effect of Geometric Initialization.}
We first compare the default and geometry-oracle settings, which differ only in the source of geometric initialization while using the same IDT-restored appearance inputs. Replacing raindrop-image geometry with GT-image geometry improves PSNR from 19.037 to 21.310 and SSIM from 0.510 to 0.620, while reducing LPIPS from 0.415 to 0.363. These gains confirm that adherent raindrops impair 3DGS not only by corrupting image appearance, but also by weakening camera pose estimation and point cloud initialization. The larger improvements in PSNR and SSIM indicate that reliable geometry is particularly important for pixel-level alignment and structural consistency. However, the geometry oracle remains far below the full oracle, showing that accurate initialization alone cannot remove the effect of raindrop-corrupted observations.

\noindent\textbf{Effect of Appearance Degradation.}
We then compare the default and appearance-oracle settings, which differ only in the input images while keeping the same raindrop-image geometry. Replacing IDT-restored images with GT images improves PSNR from 19.037 to 20.464 and SSIM from 0.510 to 0.551, and reduces LPIPS from 0.415 to 0.333. Compared with the geometry oracle, the appearance oracle produces a larger LPIPS reduction, indicating that clean appearance is especially important for perceptual fidelity. This is consistent with the visual artifacts caused by raindrop occlusion, refraction, defocus blur, and illumination distortion, which can be absorbed into the optimized Gaussians as floaters, blurred structures, or unstable appearance. Nevertheless, the modest PSNR and SSIM gains show that clean images remain constrained when poses and point clouds are still estimated from corrupted inputs.

\noindent\textbf{Coupled Effect of Appearance and Geometry.}
The full oracle, which uses both GT images and GT-image geometry, improves PSNR from 19.037 to 25.525, SSIM from 0.510 to 0.811, and LPIPS from 0.415 to 0.156. This large recovery is substantially greater than either partial oracle, indicating that appearance degradation and geometric initialization errors are complementary failure sources. 
In particular, compared with the geometry oracle, the full oracle further improves PSNR by 4.215 dB and SSIM by 0.191, while reducing LPIPS by 0.207. Compared with the appearance oracle, it further improves PSNR by 5.061 dB and SSIM by 0.260, while reducing LPIPS by 0.177. These remaining gaps across both distortion-based and perceptual metrics show that improving only restoration or only initialization is insufficient.

\subsection{Discussion}

\noindent\textbf{Implications for Raindrop-aware 3D Reconstruction.}
The results suggest that raindrop-aware 3D reconstruction should not be treated as a simple cascade of single-image restoration followed by a standard 3DGS pipeline. Real adherent raindrops perturb the input images, the geometric initialization, and the multi-view consistency assumptions used during Gaussian optimization. 
Sparse view 3DGS studies show that weak multi-view consistency and poor initialization lead to unreliable Gaussians, matching our observations under raindrop-corrupted inputs~\cite{xiao2025mcgs}.
Therefore, future systems may need to couple three components more tightly: multi-view-consistent restoration to avoid hallucinated or view-dependent textures, degradation-aware pose and point cloud estimation to down-weight unreliable raindrop regions, and uncertainty-aware 3DGS optimization to prevent transient artifacts from being absorbed into the scene representation. This direction differs from optimizing only per-image restoration quality or only making the 3D optimizer more tolerant to noisy initialization.

\noindent\textbf{Benchmark Value of Dual-focused Real-world Data.}
OmniRaindrop is designed not only to rank restoration or reconstruction methods, but also to diagnose where a raindrop-aware pipeline fails. The dual-focused acquisition protocol exposes two complementary regimes: background-focused images preserve more scene structure but contain diffuse raindrop occlusions, whereas raindrop-focused images make adherent drops sharper while severely defocusing the background. Combined with daytime and nighttime captures, aligned 2D restoration pairs, multi-view 3D scenes, rain-free references, and oracle-style geometry analysis, the benchmark allows researchers to separate appearance degradation, geometric initialization errors, and final 3DGS reconstruction failure. This diagnostic role is difficult to obtain from synthetic raindrop data or from benchmarks that evaluate only single-image removal.

\noindent\textbf{Limitations and Future Directions.}
OmniRaindrop focuses on static scenes with adherent raindrops, so it does not yet cover dynamic scenes, moving objects, rain streaks, fog, splashes, or other adverse weather factors~\cite{du2026weatherreasonseg}. The evaluated methods are representative 2D restoration, pose estimation, and 3DGS pipelines, but they do not exhaust all possible reconstruction paradigms. Future work could extend the benchmark to dynamic and multi-weather settings, incorporate temporal or cross-view supervision for restoration, and develop unified models that jointly reason about degraded appearance, unreliable geometry, and view-dependent artifacts. Despite these limitations, OmniRaindrop provides a controlled real-world foundation for studying how raindrop degradation propagates from image formation to geometric initialization and final 3D reconstruction.

\section{Conclusion}

In this paper, we introduced OmniRaindrop, a large-scale, real-world benchmark for dual-focused raindrop removal under daytime and nighttime conditions. It provides spatially aligned 2D pairs/triplets and multi-view 3D scenes covering both background-focused and raindrop-focused captures with rain-free ground truth, enabling a unified evaluation of 2D raindrop removal, camera pose estimation, point cloud initialization, and 3DGS reconstruction. 
Comprehensive experiments show that raindrop-focused scenes are particularly challenging, while nighttime scenes introduce distinct illumination-dependent degradations.
We further find that better single-image restoration does not necessarily improve 3D reconstruction, since independently restored views can introduce multi-view inconsistencies that propagate into 3DGS optimization. By characterizing these coupled effects, OmniRaindrop provides a diagnostic foundation for more robust raindrop-aware perception and has supported two consecutive NTIRE challenges on Day and Night Raindrop Removal for Dual-Focused Images.

\backmatter

\bmhead{Funding} 
This work was supported by the National Natural Science Foundation of China (Grant No. 61976017).

\bmhead{Competing interests}
The authors declare that they have no relevant financial or non-financial interests to disclose.

\bmhead{Data availability}
The OmniRaindrop dataset introduced in this work will be made publicly available upon publication.

\bibliography{sn-bibliography}

\end{document}